\documentclass{article}
\usepackage[table,xcdraw]{xcolor}
\usepackage{spconf,amsmath,graphicx}
\usepackage{diagbox}
\usepackage{multirow}
\usepackage{amsmath}
\usepackage{todonotes}
\usepackage{hyperref}

\title{Orthonormal convolutions for \\ the Rotation Based Iterative Gaussianization}
\name{Valero Laparra$^\star$ \quad Alexander Hepburn$\ddagger$ \quad J. Emmanuel Johnson$^\dagger$ \quad Jesús Malo$^\star$}
\address{$^{\star}$Image Processing Laboratory, Universitat de Valencia\\ $^{\dagger}$ Université Grenoble Alpes\\ $\ddagger$Engineering Mathematics, University of Bristol}

\begin{document}
%
\maketitle
\begin{abstract}
In this paper we elaborate an extension of rotation-based iterative Gaussianization, RBIG, which makes image Gaussianization possible. 
Although RBIG has been successfully applied to many tasks, it is limited to medium dimensionality data (on the order of a thousand dimensions). In images its application has been restricted to small image patches or isolated pixels, because rotation in RBIG is based on principal or independent component analysis and these transformations are difficult to learn and scale.
Here we present the \emph{Convolutional RBIG}: an extension that alleviates this issue by imposing that the rotation in RBIG is a convolution. 
We propose to learn convolutional rotations (i.e. orthonormal convolutions) by optimising for the reconstruction loss between the input and an approximate inverse of the transformation using the transposed convolution operation.
Additionally, we suggest different regularizers in learning these orthonormal convolutions. For example, imposing sparsity in the activations leads to a transformation that extends convolutional independent component analysis to multilayer architectures. 
We also highlight how statistical properties of the data, 
such as multivariate mutual information, can be obtained from \emph{Convolutional RBIG}. We illustrate the behavior of the transform with a simple example of texture synthesis, and analyze its properties by visualizing the stimuli that maximize the response in certain feature and layer.

\end{abstract}
\section{Introduction}
\label{sec:intro}


The rotation based iterative Gaussianization (RBIG) method \cite{Laparra09, Laparra11} relies on a sequential combination of rotation and marginal equalization transforms to arrive at a known probability density function (PDF); a zero mean unit covariance Gaussian. Since being proposed, it has been used successfully in many applications like computing different information theory measures \cite{Laparra20,Johnson2020}, analyzing remote sensing data \cite{Johnson21}, anomaly and change detection \cite{Padron22} and quantifying the information flow in the brain~\cite{Gomez19,Malo20}. However, computational restrictions have meant that applications in image processing have been infeasible. Although there are different ways to allow one to compute statistically meaningful rotations, for instance principal components analysis, these methods are computational expensive when dealing with high dimensional data like in images (more than one thousand dimensions). In deep learning this problem has been successfully alleviated by using convolutional layers instead of dense layers \cite{Deep15}. However, the constraint that the linear transformation in RBIG must be orthonormal has prevented regular convolutions from being used.

First proposed in the normalising flows literature \cite{selfnormalizing}, orthonormal convolutions are found when enforcing the transpose operation of the convolution to be the inverse. Minimizing a reconstruction error, we obtain a transformation which is nearly-orthonormal (or orthonormal if the error is zero). Then we can approximate the inverse to a convolution by applying a transpose convolution with the same filters. We propose using this operation within the RBIG framework by training an autoencoder-like structure with one forward pass (convolution) and one backward pass (transpose convolution), where the filters are shared between the layers. Unlike in normalising flows, which optimises the reconstruction error in every layer simultaneously whilst maximising the log-likelihood of the data, we obtain that a simple layer-by-layer optimization can lead to meaningful orthonormal transformations. By following the RBIG framework, we ensure that the reconstruction error for an orthonormal convolution is minimised before moving onto the next. We also propose and analyze additional regularization on the activations which maximizes the independence property of the transform, which is crucial in the RBIG framework.


\emph{Convolutional RBIG} has connections with many other fields. 
On the one hand, as shown below, it extends the convolutional independent components analysis proposed in \cite{Balle2014}. While the (single layer) solution in \cite{Balle2014} has nice statistical properties, it is computationally expensive which makes it infeasible to be used in the RBIG framework, where many of these transformations are performed in a multilayer architecture. On the other hand, \emph{Convolutional RBIG} is also closely related to the field of normalizing flows \cite{Flows}, as pointed out 
in~\cite{Johnson2020}. 
The restrictions in normalizing flows led to the use of convolutions with a spatial filter size of 1, 
which is a strong constraint for images. 
Finally, \emph{Convolutional RBIG} is also closely related to statistical models of images.
In~\cite{balle2015density}, convolutions followed by parametric and multidimensional nonlinearities have been used to obtain a statistical model of the natural images. However this makes the learning process much more complex. Here we restrict to the original idea of using simple and easy to compute transformations, learn the transformations layer by layer, and show that \emph{Convolutional RBIG} can also be used to obtain statistical descriptors of the images used for training.


\section{Convolutional RBIG}
\label{sec:OrthoConv}

The original RBIG~\cite{Laparra11} finds a transform, $G({\bf x})$, that converts the input data to a domain where it follows a standard normal distribution, $G({\bf x}) \sim \mathcal{N}(0,I)$. The transform $G({\bf x})$ consists of several layers, where each layer is defined as the concatenation of two operations: a marginal Gaussianization $\Psi(\cdot)$, and a rotation $R$, i.e. one RBIG layer is ${\bf x}_{i+1} = L_{i}({\bf x}_{i}) = R \cdot \Psi({\bf x}_{i})$. Therefore an RBIG transform of $P$ layers is:
\begin{equation}
G({\bf x}) = L_{P} \left( \textdegree L_{P-1} \left(  \textdegree ... L_{1}({\bf x})  ... \right) \right).
\label{eq:RBIG}
\end{equation}
\subsection{Orthonormal Convolutions.}

The core of the proposed \emph{Convolutional RBIG} is using orthonormal convolutions in the rotation step, $R$, as opposed to the generic rotations in the original RBIG. This simple idea is easy to implement and does not require extensive training.

To be a rotation, a linear transform must keep the dimensionality and be composed of orthonormal vectors. 
In the linear transforms we are aiming at (convolutions) the number of filters must be restricted to preserve dimensionality. We define the $i$-th convolutional filter as a tensor, $c_{i}$, with size $l_1 \times l_2 \times ch_{in} \times ch_{out}$, where $l_1$ and $l_2$ are the kernel $height$ and $width$ respectively and $ch_{in}$ and $ch_{out}$ are the number of channels of the input and the output, i.e. an RGB image has 3 channels. As seen later, one can use more than one filter in our proposal for orthonormal convolution (i.e. $i>1$), but for now we restrict ourselves to a single filter for simplicity. 

An image, $X$, is a tensor of size $h\times w\times ch$, and its vectorized version, ${\bf x}$, is a vector of size $h \cdot w \cdot ch \times 1$, where $h$ and $w$ are the height and width of the image and $ch$ is the number of channels. While the filter sizes do not depend on $h$ and $w$ of the input, the associated convolution matrix, $C$, derived from the filter, $c$, will depend on them: $C$ has size $ h \cdot w \cdot ch_{out} \times h \cdot w \cdot ch_{in}$. Therefore these two operations are equivalent, $c * X \equiv C \cdot {\bf x}$. 
On the other hand the transpose convolution operation, $\tilde{*}$, is equivalent to apply the transpose of the convolution matrix over the vector: $c \tilde{*} X \equiv C^\top \cdot {\bf x}$. 

Our idea is based on defining this mini-autoencoder:
\begin{equation}
    \hat{\bf x} = C^\top \cdot C \cdot {\bf x}
\label{eq:autoencoder}
\end{equation}

By training this mini-autoencoder to minimize the reconstruction error, $||{\bf x}-\hat{{\bf x}}||_2$,  we enforce the convolution matrix to be orthonormal, i.e. $C^\top C = I$. Note that we are using the same convolution filters for both the forward (convolution) and the backward (transpose convolution) pass. Ideally the reconstruction error would be zero, and therefore the transformation would be orthonormal. This is possible without imposing any other restriction. In Sec.~\ref{sec:regularization} we will explore some restrictions that can help on the Gaussianization step while minimally penalizing the reconstruction error restriction. 

The normalizing flow literature~\cite{GLOWKingmaD18} introduced 
$1\times1$~convolutions parameterized by an LU decompositions and random permutations. These were useful but there were some stability issues during training due to the random permutations. This was improved in \cite{emergeconv} where they enforced the kernel in the $1\times1$~convolution to be orthogonal. These $1\times1$~convolutions do not account for the spatial variability without shuffling and reshaping the dimensions of the image~\cite{realnvp,nice}. In addition, 
the same authors~\cite{emergeconv} created convolutions which accounted for spatial domain via an autoregressive transformation however this was very expensive to invert. Circular convolutions~\cite{invconvflow,invconvnet} that assume periodic boundaries have also been used but this is not a good inductive bias for many types of images.
The authors in \cite{selfnormalizing} perform an operation very similar to the method introduced above. However, their proposal is used as a term in the whole optimization process, while here we propose its optimization layer by layer, enforcing gaussianize orthonormality. In addition, an extra regularization term is proposed to accelerate convergence which minimizes the number of layers required. 

\subsection{Learning and regularization}
\label{sec:regularization}

While in the original proposal~\cite{Laparra11} three different types of rotations were analyzed, here we propose two possibilities for extra constrains in the learning process which bring to four different solutions. This constrains will be added in learning each mini-autoencoder, $C$, in Eq.\ref{eq:autoencoder}. 

The first regularization strategy consists in enforcing sparsity in the inner domain, i.e penalizing the activity after the convolution $||C {\bf x}||_1$. This is the principle behind most independent component analysis algorithms. It enforces the convolution to convert the data to a domain where the marginals have less entropy. And therefore increasing the impact of $\Psi$ in converting the data to a standard normal. Its importance is modulated by the hyperparameter $\lambda_{act}$. The second proposed regularization (modulated by $\lambda_{c}$) is the classical one over the convolution weights values, $||c||_1$ ,which bring to cleaner convolution kernels. Therefore mini-autoencoder cost function is:  
\begin{equation}
    \mathcal{L}_c = ||{\bf x}-\hat{{\bf x}}||_2 + \lambda_{act} ||C \cdot {\bf x}||_1 +  \lambda_{c} ||c||_1 
\label{eq:cost_function}
\end{equation}

\begin{figure}[h]
\begin{tabular}{c}
\hspace{-0.5cm}\includegraphics[width=9cm,height=4.5cm]{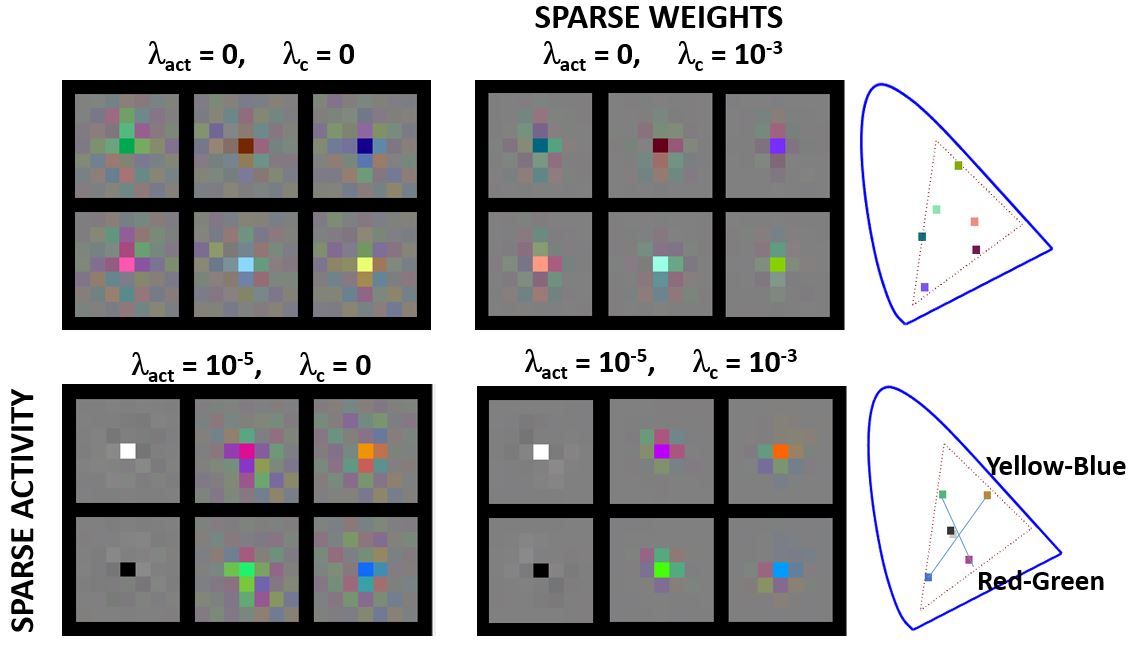}
\end{tabular}
\vspace{-0.4cm}
\caption{Filters obtained in the first layer using stride 1. We used filter size $7\times7$ and different combinations of regularization, see Eq.\ref{eq:cost_function}. 
Positive and negative versions of the filters are shown at the top and bottom in each panel. Enforcing sparse activity leads to center-surround filters with clear red-green / yellow-blue opponent directions.}
\label{fig:FILTERS_stride_1_FW_7_regularization}
\end{figure}

Figure \ref{fig:FILTERS_stride_1_FW_7_regularization} shows 
the positive effects of regularizations:
on the one hand, enforcing the sparsity of the weights leads to cleaner (i.e. less noise artifacts) convolutional kernels.
And more interestingly, sparsity on the activity when gaussianizing natural images leads to the achromatic, red-green, yellow-blue center surround kernels that have been found when imposing optimal information transmission~\cite{Atick92b,Li2021}. Next section shows the effect of these regularizers on the learning process and the statistical properties of the transformation.

\subsection{Computing multi-information reduction.}
\label{ssec:comp_MI}

If we assume perfect reconstruction, the transformation would be orthonormal and therefore by definition the determinant of the convolution matrix would be one, i.e. $|C| = 1$. As it was shown in~\cite{Laparra11} this allows to analyze some properties of the original data. Here we highlight the most interesting one.  

\begin{figure}[t]
\centering
\begin{tabular}{c}
\includegraphics[width=7cm]{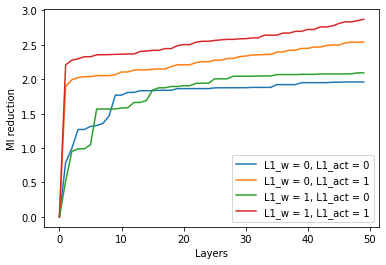}
\end{tabular}
\vspace{-0.4cm}
\caption{Accumulated MI reduction on validation with different combinations of regularization.}
\label{fig:Acc_MI_stride_1_FW_7_regularization}
\end{figure}

Following the reasoning in~\cite{Laparra11} the marginal gaussianization layers, $\Psi$, do not reduce the multivariate multi-information (MI), while the amount of MI in the signal reduced by the linear layers, ${\bf y}=C\cdot {\bf x}$, is:
\begin{equation}
    \Delta MI = \\
    \sum_{d=1}^D h( x_d ) - \sum_{d=1}^D h(y_d) + log(|C|) 
\end{equation}

\noindent where $h(\cdot)$ is the univariate differential entropy and $D$ is the data dimension. 
In the case of rotations the last term vanishes, and $\Delta MI$ only depends on marginal computations (see \cite{Laparra20} for details). 
We will use accumulated MI reduction to illustrate how much improvement a new layer introduces in the learning process. Its interpretation is very similar to the log likelihood. For instance Fig. \ref{fig:Acc_MI_stride_1_FW_7_regularization} shows the effect in mutual information reduction when stacking multiple layers. It is clear that the method reduces the mutual information during training (as expected).
This plot also illustrates the effect of the different regularizations in the MI reduction. While it is clear that the biggest (positive) effect comes from using the activity regularization, it is also clear the positive effect of using the regularization over the weights.

\subsection{Learning multiple filters}
\label{ssec:multipleFilt}

While using the same number of filters as input channels naturally enforces the requirement of keeping the dimensionality constant, this might not be ideal for some applications. For instance the filters in Fig.\ref{fig:FILTERS_stride_1_FW_7_regularization} show a kind of center surround architecture as expected in first stages of an artificial neural network that deals with images, but this strong restriction does not allow the method to find more complex filters like wavelets. In order to give more flexibility to the method but keeping the dimensionality constant, one can subsample the spatial dimension and increase the number of filters/channels. 

In order to prevent a collapse of the spatial dimension before gaussianizing the data, the undersampling-layers can be combined with layers that 
keep the spatial dimension. 
This flexibility allows richer models that may be better suited for images. Specifically, our strategy consists of using "blocks" composed by one layer that performs a subsampling (i.e., stride $>1$) followed by several layers that do not (i.e., stride $1$). 
For instance Fig. \ref{fig:Text_grass} 
illustrates the effect of using an architecture with 5 blocks (with stride = 2) and 5 layers in each block. Note that the introduction of the layers that reduce the dimensionality has a clear positive effect in mutual information reduction. An example of the filters obtained in the first reduction layer is shown in Fig.~\ref{fig:FILTERS_stride_4_FW_12} (same idea as in Fig.~\ref{fig:FILTERS_stride_1_FW_7_regularization} but using stride 4 instead).

\section{Experiments}
\label{sec:results}
Here we show a simple example of texture synthesis, and the effect when different amounts of layers are used. We also show an analysis of the transformation inferred by the model. 
All model results in this paper have been trained using the cifar10~\cite{cifar10} dataset except those in the texture section. Code example of the basic modules of the proposal can be found in \url{https://github.com/alexhepburn/ortho\textunderscore conv}.

\subsection{Texture Synthesis}
\label{ssec:Exp_synthesis}

\begin{figure}[htb]
\centering
\begin{tabular}{c}
\includegraphics[width=7.5cm]{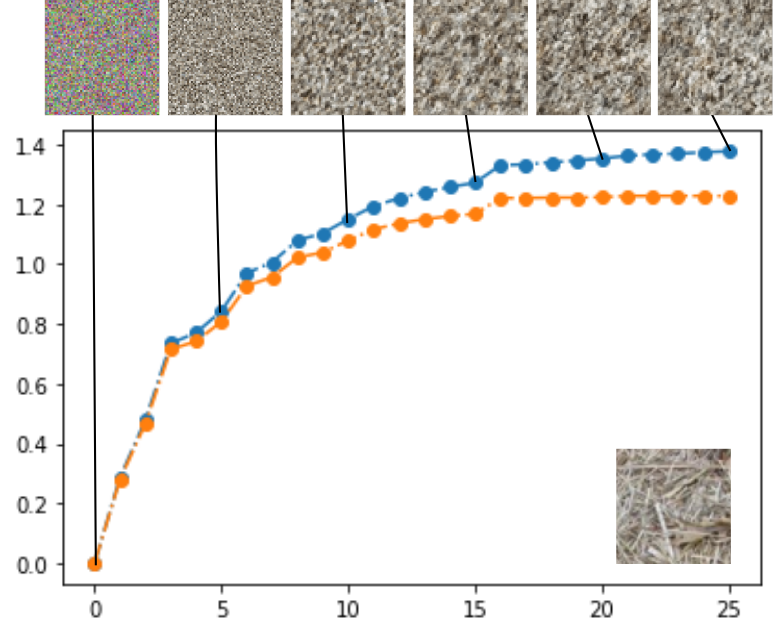}
\end{tabular}
\vspace{-0.5cm}
\caption{Accumulated MI reduction on train (blue) and validation (orange) with a synthesized image corresponding to inverting a Gaussian noise sample after the output of each block. The original texture is shown over the plot.}
\label{fig:Text_grass}
\end{figure}

The proposal has several advantages. One of its strengths is that it does not require hyperparameter adjustment in addition to deciding the number of layers. And, in fact, the reduction of mutual information acts as a criterion for deciding whether adding more layers will make any improvement. Another advantage is that it can easily be used as a texture synthesizer. It is only necessary to train the model on a particular texture, generate random noise (following a standard Gaussian distribution) and invert the data using the trained model. Figure \ref{fig:Text_grass} shows an example of synthesis for a particular texture. The example illustrates the two aforementioned facts. On the one hand it is easy to generate textures form the model for different configurations. On the other hand it is clear how the amount of MI reduction achieved by the model corresponds to quality in the synthesized texture. Therefore the MI reduction can be used as a criteria to decide the model complexity.   

\vspace{-0.3cm}
\subsection{Analysis of the representation}
\label{ssec:Exp_representation}

\begin{figure}[htb]
\begin{center}
\includegraphics[width=8cm,height=3.6cm]{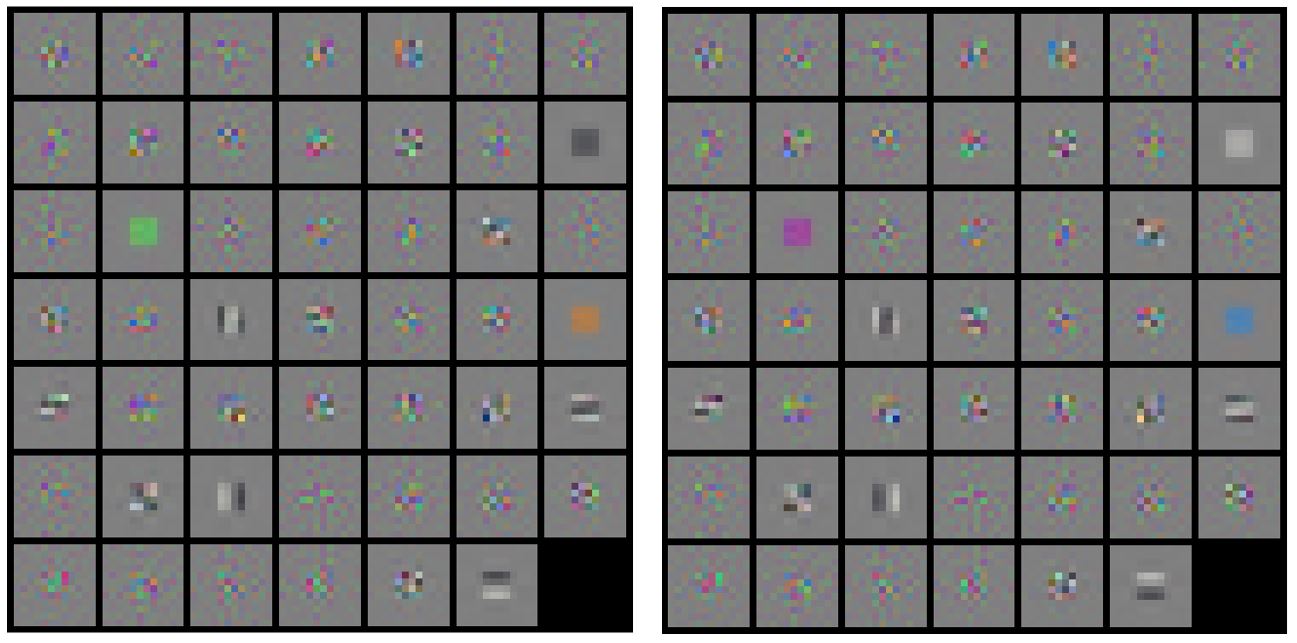}
\end{center}
\vspace{-0.7cm}
\caption{Learning multiple filters allows the emergence of image features beyond center-surround. 
Here filters obtained in the first convolutional layer using stride 4 and sparse activity.}
\label{fig:FILTERS_stride_4_FW_12}
\end{figure}

In Fig.~\ref{fig:FILTERS_stride_1_FW_7_regularization} we showed the 1st-layer result of the convolutional-RBIG strategy in natural images when using stride 1. In Fig.~\ref{fig:FILTERS_stride_4_FW_12} show the same results but using multiple filters strategy (Sect.~\ref{ssec:multipleFilt}), in this case we use stride = 4. 
Bigger strides allow the emergence of additional convolutional-orthogonal features different from center-surround patterns. Localized oriented receptive fields in opponent chromatic directions is consistent with the literature on ICA in natural color images~\cite{Doi03,Hyvrinen2009,Gutmann14}.
However, as opposed to these classical results, our receptive fields are convolutional (as in~\cite{Balle2014}), with the additional advantage that further layers in convolutional-RBIG capture bigger fraction of the information in the signal (as illustrated in Figs.~\ref{fig:Acc_MI_stride_1_FW_7_regularization},\ref{fig:Text_grass}).

\begin{figure}[htb]
\begin{center}
\includegraphics[width=8.3cm,height=6.2cm]{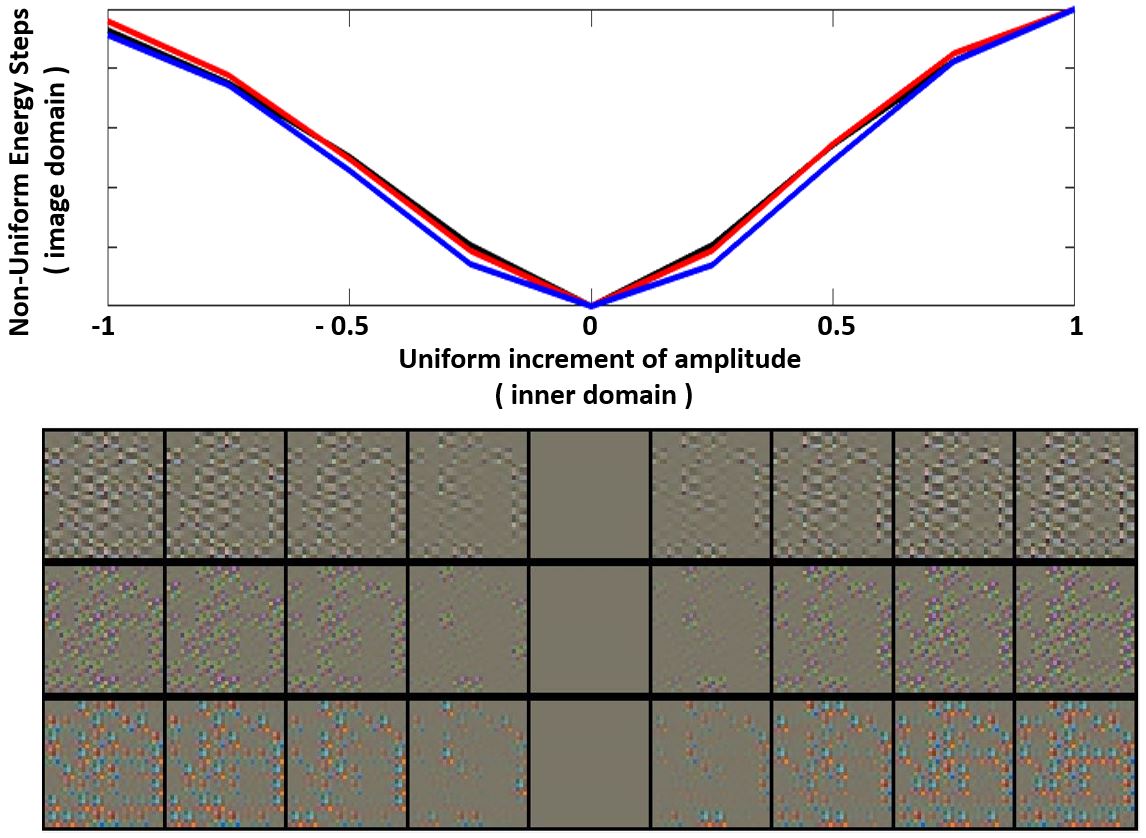}
\end{center}
\vspace{-0,5cm}
\caption{Uniform displacements in the cardinal directions of the inner representation lead to nonuniform variations of the energy of the texture that induces these variations.}
\label{Fig_nonlinearity}
\end{figure}

The invertible nature of the convolutional RBIG simplifies the visualization procedures required in other networks for response analysis~\cite{Vedaldi14}. In Fig.~\ref{Fig_nonlinearity} we explore three particular directions of the inner representation obtained after the marginal gaussianization that follows the filters in Fig.~\ref{fig:FILTERS_stride_4_FW_12}. Starting from the images that elicit zero response (flat gray images), we inject white noise of increasing amplitude in bands corresponding to achromatic, RG and YB textures and compute the inverse. The observerd saturation of the output back in the image space implies a contrast-dependent resolution in the inner space.  

\vspace{-0.4cm}
\section{Conclusion}
\label{sec:conclusion}

Here we present a simple way to obtain orthonormal (or quasi-orthonormal) convolutions. It is implemented in the framework of rotation-based iterative gaussianization (RBIG) in order to make it capable of dealing with images efficiently. The proposal has several advantages inherited from the original RBIG (easy to train, no need to adjust hyperparameters, allows to compute statistical properties of the data, is invertible, easy to compute the Jacobian...) and adds the ability to deal naturally with images. We show different options (such as different regularization terms, and different possibilities in the architecture) and properties, and illustrate them in different experiments. This work opens up multiple options for the use of the proposed new architecture that encourage further exploration.    


\bibliographystyle{IEEEbib}
\bibliography{strings,refs}

\newpage

\end{document}